\documentclass{osa-article}
\journal{osajournal}
\articletype{Research Article}
\usepackage{amsmath}
\usepackage{subfig}
\usepackage{graphicx} 
\usepackage{pdfpages}
\usepackage{hyperref}

\usepackage[utf8]{inputenc}
\usepackage[english]{babel}

\ifx\figurename\undefined \def\figurename{Figure}\fi
\renewcommand{\figurename}{Fig.}
\renewcommand{\paragraph}[1]{\textbf{#1} }

\newcommand{\Sect}[1]{Sec.~\ref{#1}}
\newcommand{\Fig}[1]{Fig.~\ref{#1}}

\newcommand{\Equ}[1]{Eq.~(\ref{#1})}

\newcommand{\RNum}[1]{\uppercase\expandafter{\romannumeral #1\relax}}

\newcommand{\sbl}[1]{_{\mathrm{#1}}}

\begin{document}
\title{A Design Framework for Metasurface Optics-based Convolutional Neural Networks\footnote{\small This paper can be accessed from the Optical Society of America's website, at\\ \url{https://www.osapublishing.org/ao/abstract.cfm?uri=ao-60-15-4356}}}

\author{Carlos Mauricio Villegas Burgos,\authormark{1} Tianqi Yang,\authormark{2} Yuhao Zhu,\authormark{2,3} and A. Nickolas Vamivakas\authormark{1,*}}

\address{\authormark{1}University of Rochester, Institute of Optics, 275 Hutchison Road, Rochester, NY 14627, USA\\
\authormark{2}University of Rochester, Department of Computer Science, 2513 Wegmans Hall, Rochester, NY 14627, USA\\
\authormark{3}yzhu@rochester.edu}

\email{\authormark{*}nick.vamivakas@rochester.edu}

\graphicspath{{figs/}}


\begin{abstract}
Deep learning using convolutional neural networks (CNNs) has been shown to significantly out-perform many conventional vision algorithms. Despite efforts to increase the CNN efficiency both algorithmically and with specialized hardware, deep learning remains difficult to deploy in resource-constrained environments. In this paper, we propose an end-to-end framework to explore how to optically compute the CNNs in free-space, much like a computational camera. Compared to existing free-space optics-based approaches which are limited to processing single-channel (i.e., grayscale) inputs, we propose the first general approach, based on nanoscale metasurface optics, that can process RGB input data. Our system achieves up to an order of magnitude energy saving, simplifies the sensor design, all the while sacrificing little network accuracy.
\end{abstract}

\section{Introduction}
\label{sec:intro}

Deep learning (a.k.a., deep neural networks, or DNNs) have become instrumental in a wide variety of tasks ranging from natural language processing~\cite{mikolov2010recurrent, mikolov2013distributed, sutskever2014sequence} and computer vision~\cite{redmon2016you, bertinetto2016fully} to drug discovery~\cite{chen2018rise}. However, despite enormous efforts both algorithmically~\cite{lecun1990optimal, han2015deep} and with specialized hardware~\cite{jouppi2017datacenter, chen2016eyeriss}, deep learning remains difficult to deploy in resource-constrained systems due to the tight energy budgets and computational resources. The plateau of current digital semiconductor technology scaling further limits the efficiency improvements, and necessitates new computing paradigms that could move beyond current semiconductor technology.

Optical deep learning has recently gained attention because it has the potential to process DNNs efficiently with low energy consumption. A lens executes the Fourier transform ``for free''~\cite{goodman2005introduction}. Leveraging this property, one can custom-design an optical system to execute convolution (or matrix multiplication)~\cite{farhat1985optical, psaltis1988adaptive, lu1989two, saxena1995adaptive}, since convolution can be performed with the help of Fourier transform operations. The effective convolution kernel is given by the optical system's point spread function (PSF).

\paragraph{Motivation and Challenge} Existing free-space optical mechanisms can not handle layers with multiple input channels~\cite{chang2018hybrid, lin2018all, bueno2018reinforcement, AmpOnly4fCNN}, which limits the applicability to a single layer with one single input channel, e.g., MNIST hand-written digits classification. Fundamentally, 
this limitation is because their systems could not provide different and independent PSFs for multiple frequencies of light, as they use elements such as digital micromirror devices (DMDs)~\cite{DMDs} or diffractive optical elements (DOEs)~\cite{o2004diffractive} to engineer their system's PSF. DMDs modulate the amplitude of light, but impart the same amplitude modulation to all frequencies of light. Similarly, DOEs modulates the phase of incident light by imparting a propagation phase, but the phase profile imparted to one frequency component is linearly correlated with the phase profile imparted to another frequency component. Thus, these elements can not provide different and independent PSFs for multiple frequencies, and are limited to only single-channel convolution kernels.

\paragraph{Our Solution} To enable generic convolution with multiple input channels and widen the applicability of optical deep learning, we propose to use \textit{metasurface}~\cite{meta-review, meta-theory} optics, which are ultra-thin optical elements that can produce abrupt changes on the amplitude, phase, and polarization of incident light. Due to the flexibility to shape light properties along with their ultra-compact form factor, metasurfaces are increasingly used to implement a wide range of optical functionalities. 
One such functionality, which we use on our work, is imparting a geometrical phase to circularly polarized light that interacts with the nano-structures (referred to as \textit{meta-elements} from here on out) on the metasurface. The portion of incident light that gets embued with this geometrical phase also has its handedness changed \cite{meta-review, MetaPol}. The power ratio between light embued with this geometrical phase and the total incident light is a function of frequency, and this function is referred to as the polarization conversion efficiency (PCE) \cite{wang2016visible}.

\begin{figure}[t]
\centering
\captionsetup{width=.9\columnwidth}
\includegraphics[width=.9\columnwidth]{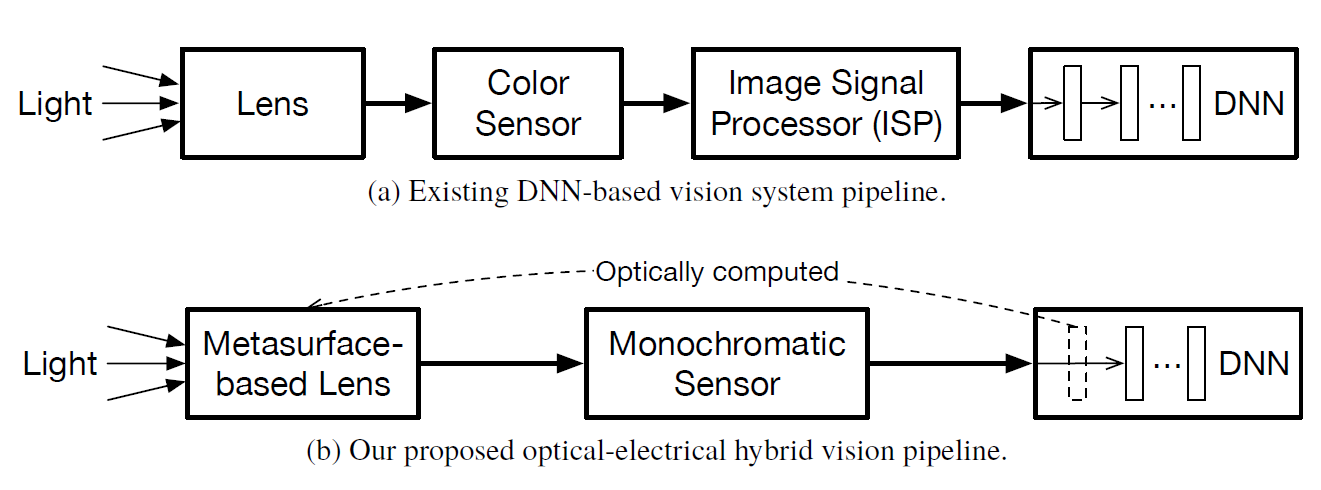}
\caption{Comparing our proposed system with conventional systems. Our system moves the first layer of a DNN into the optical domain based on novel metasurface-based designs. This hybrid system allows us to use a simple monochromatic sensor while completely eliminating the image signal processor (ISP), which is power-hungry and on the critical path of the traditional pipeline.}
\label{fig:overview}
\end{figure}

Given a target convolution layer to be implemented optically, we purposely design the metasurface such that each of the meta-elements in a given unit cell has a unique PCE spectrum that responds to only one specific, very narrow frequency band corresponding to a particular input channel. In this way, metasurfaces let us apply different phase profiles to different input channels, which enables the system to have different PSFs for each channel, which effectively allows processing multi-channel inputs.

However, using metasurface optics for optical deep learning is not without challenges. Determining the metasurface parameters to produce a given PCE specification, known as the inverse design problem of a metasurface, is known to be extremely time-consuming, because it involves exploring billions of design points, each of which requires simulating the time evolution of the system by computing solutions to Maxwell's equations~\cite{pestourie2018inverse, molesky2018inverse}. Second, an optical layer when directly plugged into a DNN model will likely not match the exact results of the mathematical convolution due to optical design imperfections and sensor artifacts (e.g., noise).

We propose an \textit{end-to-end} optimization framework that avoids the inverse design problem altogether while minimizing the overall network loss. We start from existing meta-element designs whose PCEs are close enough to our desired PCE specifications. We then accurately model the metasurface behaviors to integrate the metasurface into an end-to-end design framework, which co-optimizes the metasurface design with the rest of the network including the analog sensor and the digital layers in a DNN model. Critically, the end-to-end pipeline is differentiable, which allows us to generate a complete system design (metasurface optical parameters and the digital layer parameters) using classic gradient descent methods.

\paragraph{Contributions} Leveraging the end-to-end design framework, we explore a system where the first layer in a DNN is moved to the optical domain; the image sensor captures the convolutional layer's output feature map, which is then processed in the digital domain to finish the entire network. \Fig{fig:overview} illustrates the idea of our system.

Our design has three advantages. First, since the sensor captures the feature map rather than the original RGB image, our system enables us to use a simple monochromatic sensor rather than a RGB sensor used in conventional systems, eliminating engineering efforts of many components such as the Color Filter Array (CFA).

Second, since we use a monochromatic sensor that captures the output feature map, we can avoid the often power-hungry computations in the image signal processor (ISP), which is required by conventional RGB sensors to correct sensor artifacts such as demosaicing and tone mapping. Avoiding ISP leads to significant overall energy reduction.

Finally, by moving the first layer into optics, we reduce the amount of computation left in the digital domain.

In summary, this paper makes the following contributions:

\begin{itemize}
	\item We are the first to address generic (multi-channel) convolution in free-space optics. This is achieved by a novel use of metasurface optics to independently modulate the phase that is imparted to different frequency components of incident light.
	\item We present an end-to-end differentiable optimization framework, which allows us to co-train the optical parameters and digital layer parameters to maximize the overall accuracy and efficiency.
	\item We evaluate our system on multiple vision networks (AlexNet, VGG, and ResNet50). Our simulation results show that our system can reduce the overall system energy by up to an order of magnitude while sacrificing an accuracy loss of 1.9\% on average (0.8\% minimum).
\end{itemize}

\section{Related Work}
\label{sec:related}

\paragraph{Integrated Photonics DNN} Existing optical deep learning systems take two major approaches: integrated photonics~\cite{shen2017deep, xia2007ultra, bagherian2018chip} and free-space optics~\cite{lin2018all, chang2018hybrid, bueno2018reinforcement, hamerly2019large}. Burgos et al.~\cite{burgos2019challenges} propose a Multi-Layer Perceptron (MLP) implementation based on microring resonators (MRRs)~\cite{bogaerts2012silicon}. Other implementations used different optical mechanisms such as Mach-Zehnder interferometers (MZI)~\cite{shen2017deep, bagherian2018chip}, but the general principle holds. Integrated photonics circuits are fundamentally designed to execute matrix multiplications. Convolutions must be first converted to matrix multiplication, and then mapped to the photonics substrate~\cite{bagherian2018chip}.

The advantages of integrated photonics is that it could be directly integrated into CMOS chips, to minimize the form-factor. Integrated photonics could potentially enable optical training~\cite{hughes2018training}. The disadvantage of integrated photonics is two-fold. First, they fundamentally require coherent light (generated from a laser source) and thus could not directly operate on real-world scenes, where light is incoherent. Second, due to the quadratic propagation losses of integrated waveguides, the power consumption of the integrated photonics system could be extremely high when implementing realistic DNN models~\cite{burgos2019challenges}. The power goes up to 500~W to execute an MLP layer with $10^5$ neurons, which in turn limits the scale of the DNN to be implemented.

\paragraph{Free-Space Optical DNNs} Motivated by the limitations of integrated photonics, recent research starts exploring free-space optics, which exploit the physical propagation of light in free space through custom lens to perform convolution. Chang et al.~\cite{chang2018hybrid} demonstrate a 4-$f$ system using DOEs to implement convolution layers, which support multiple output channels but does not support multiple input channels. Lin et al.~\cite{lin2018all} proposes a DOE-based free-space design, which supports a unique network architecture that is neither an MLP nor a CNN: it consists of the basic structures of an MLP, but has a limited receptive field typically found in CNNs. Miscuglio et al.~\cite{AmpOnly4fCNN} demonstrate a 4-$f$ system where a DMD is placed in the Fourier plane to implement the convolutional kernels of a CNN layer by modulating only the amplitude of the incident light. However, this system is also limited to only being able to process single-channel inputs.

\paragraph{Deep (Computational) Optics} Implementing deep learning (convolution) in free-space optics is reminiscent of classic computational optics and computational photography, which co-designs optics with image processing and vision algorithms~\cite{zhou2011computational}. Recent progresses in deep learning has pushed this trend even further to co-train optical parameters and DNN-based vision algorithms in an end-to-end fashion. Examples include monocular depth estimation~\cite{chang2019deep, haim2018depth, he2018learning, wu2019phasecam3d}, single-short HDR imaging~\cite{sun2020learning, metzler2019deep}, object detection~\cite{chang2019deep}, super-resolution and extended depth of field~\cite{sitzmann2018end}.

\paragraph{Computing in Metasurface Optics} Lately, metasurface optics have been used for various computation tasks through co-optimizing optics hardware with backend algorithms. Lin et al. \cite{lin2020end} uses a simple ridge regression as an example backend while our work employs a neural network. Lin et al. focuses on exploiting fullwave Maxwell physics in small multi-layer 3D devices for non-paraxial regime, and enhanced spatial dispersion, spectral dispersion and polarization coherence extraction. In contrast, our work focuses on exploiting large-area single-layer metasurfaces for computer vision tasks.

Colburn et al. \cite{colburn2019optical} uses metasurface for convolution, too. To support multiple input channels, however, they use three separate laser sources, each going through a separate 4-$f$ system, whereas we leverage polarization to support multiple input channels with one 4-$f$ system, potentially simplifying the design and the overall form factor of the system.

\section{Fundamentals of Free-Space Optics}
\label{sec:fs}

A typical setup for free-space optical DNN is similar to computational cameras~\cite{zhou2011computational, ng2005light}, where a custom-designed lens directly interrogates the real-world scene as an optical input signal. The lens executes the first convolution or fully connected layer of a DNN; the layer output is captured by an image sensor and sent to the digital system to execute the rest of the DNN model. This setup is a natural fit for machine vision applications where the inputs are already in the optical domain and have to be captured by a camera sensor anyways.

Free space imaging scenarios can be modeled by assuming spatially incoherent light and a linear, shift-invariant optical system. Under these assumptions the image formed by an optical system ($I_{out}$) can be modeled as a convolution of the scene, i.e., input signal ($I_{in}$), with the point spread function (PSF) of the optical system~\cite{goodman2005introduction}:

\begin{equation}
\label{eq:conv}
I_{out}(x, y, \omega) = I_{in}(\eta, \xi, \omega) \star \mathbf{PSF}(\eta, \xi)
\end{equation}

\noindent where $\star$ denotes convolution, $I_{in}(\eta, \xi, \omega)$ denotes the intensity of input pixel ($\eta$, $\xi$) at frequency $\omega$, and $I_{out}(x, y, \omega)$ denotes the intensity at output pixel ($x$, $y$) at frequency $\omega$.

\begin{figure}[t]
  \centering
  \includegraphics[width=.7\columnwidth]{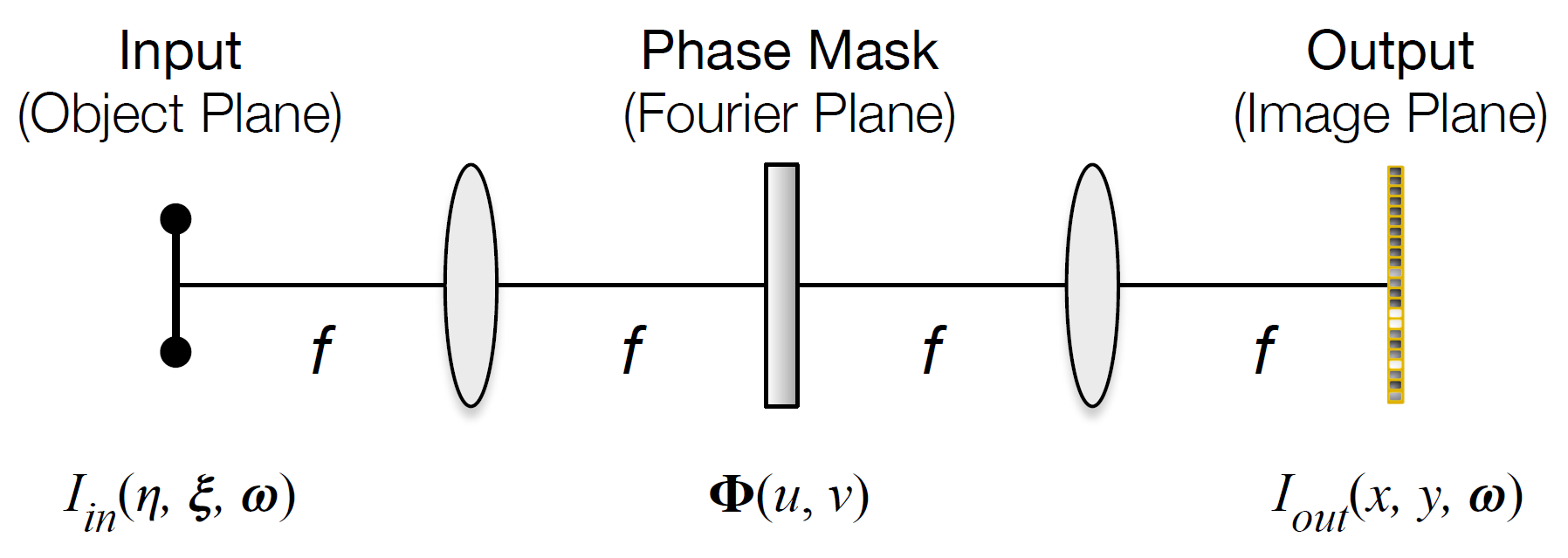}
  \caption{A typical ``4-$f$'' system.}
  \label{fig:4f}
\end{figure}

An optical convolution system is typically implemented using a so-called ``4-$f$'' system~\cite{goodman2005introduction} as illustrated in~\Fig{fig:4f}, which consists of two lenses, each with focal length $f$. The first lens is placed one $f$ away from the object plane, producing a Fourier plane another $f$ after the first lens. The second lens is placed another $f$ from the Fourier plane, producing an inverted conjugate image plane, which is in total 4-$f$ from the original object plane. An image sensor is placed at the image plane to capture the output for future processing (e.g., activation and subsequent layers) in the electronic domain. The reason why the back focal plane of the first lens is referred to as Fourier plane is because the electric field on that plane can be expressed in terms of the Fourier transform of the electric field on the front focal plane. Mathematically speaking, let $U_{i}(\eta,\,\xi)$ be the field in the input plane of the 4f system, where $(\eta,\,\xi)$ denote the spatial coordinates in that plane. Also, let $U_{F}(u,\,v)$ be the field in the Fourier plane of the 4-$f$ system, where $(u,\,v)$ denote the spatial coordinates in that plane. Then, the field in the Fourier plane is expressed in terms of the field in the input plane as:
\begin{equation}
\label{eq:FourierPlane}
U_{F}(u,\,v) = F_{i}\left(\frac{u}{\lambda f},\,\frac{v}{\lambda f}\right),
\end{equation}

\noindent where $\lambda$ is the wavelength of the incident light, $f$ is the focal length of the lenses in the 4-$f$ system, and $F_{i}(f_{\eta},\,f_{\xi})$ is the Fourier transform of $U_{i}(\eta,\,\xi)$, with $(f_{\eta},\,f_{\xi})$ denoting spatial frequency components.

The PSF of the optical system is dictated by the \textit{phase mask} placed at the Fourier plane. The phase mask alters the PSF by imparting certain phase change to the incident light. It achieves this by introducing a phase term to the field on the Fourier plane, meaning it multiplies $U_{F}(u,\,v)$ by $\mathrm{exp}\left(i\Phi(u,\,v)\right)$ (here the imaginary unit is denoted as $i$) .The phase change induced by a phase mask is called the phase mask's \textit{phase profile}, which is often determined by the geometric properties of the meta-elements on the phase mask (e.g., dimension, orientation, thickness). Our goal is to find the phase profile $\Phi$ to generate a PSF that matches a target image that contains the intended convolution kernel $W$. This can be formulated as an optimization problem:

\begin{equation}
\label{equ:opt}
\min_{\Phi}\vert\vert\, W - \mathbf{PSF}(\Phi)\vert\vert.
\end{equation}

Without the phase mask, the image captured by the image sensor simply reproduces the input $I_{in}$. With the optimized phase mask, the output image is naturally the result of the convolution in \Equ{eq:conv}.

The output of a generic convolutional layer is given by:

\begin{equation}
\label{eq:cnn}
I_{out}^{j}(x, y) = \sum_{i=1}^{C_{in}} I_{in}^{i}(x, y) \star W_{j,i}(x, y),~~~~\forall j \in [1, C_{out}].
\end{equation}

\noindent The optical convolution in \Equ{eq:conv} can be seen as a simplification of \Equ{eq:cnn} whenever the number of input channels ($C_{in}$) and the number of output channels ($C_{out}$) of the convolutional layer are both equal to 1. Supporting multiple output channels (i.e., multiple convolution kernels) is achieved by designing a PSF that spatially tiles the $C_{out}$ convolution kernels on the same plane; each tile corresponds to a convolution kernel $W_{j, \cdot}$~\cite{chang2018hybrid}. As the input image is convolved with the system's PSF, it is naturally convolved with the different convolution kernels, effectively generating multiple output channels.

However, supporting multiple input channels is a fundamental challenge unaddressed in prior work. Essentially, the optical system must have multiple PSFs each of which must apply to \textit{only a particular frequency} (i.e., channel) of the incident light. This is inherently unattainable in prior works that use diffractive optical elements (DOE)~\cite{o2004diffractive} to implement the phase mask. This is because the phase profile that is implemented for one frequency is a scalar multiple of the one that is implemented for other frequencies. As such, it is not possible to create independent phase profiles for the different frequency components of incident light.
To overcome this limitation, we instead propose to implement the phase mask using geometrical phase metasurfaces, which we describe in detail next.

\section{Metasurface-based Optical Convolution}
\label{sec:sys:ms}

The goal is to design the phase mask in a way that different PSFs are applied to different input channels. Unlike DOEs that modulate only the phase of the incident light by applying a propagation phase, metasurfaces are able to also introduce a geometrical phase. Our key idea is to use different kinds of meta-elements that embue a geometrical phase to incident light of different frequency components. That way, the metasurface introduces different phase profiles to each of the frequency components (channels) of incident light.
In the following, we first describe the phase modulation assuming only one input channel; and then discuss the case with multiple input channels, which allows us to attain generalized free-space optical convolution with multiple input and output channels.

\begin{figure*}[t]
  \centering
  \captionsetup{width=\columnwidth}
  \includegraphics[width=1.0\columnwidth]{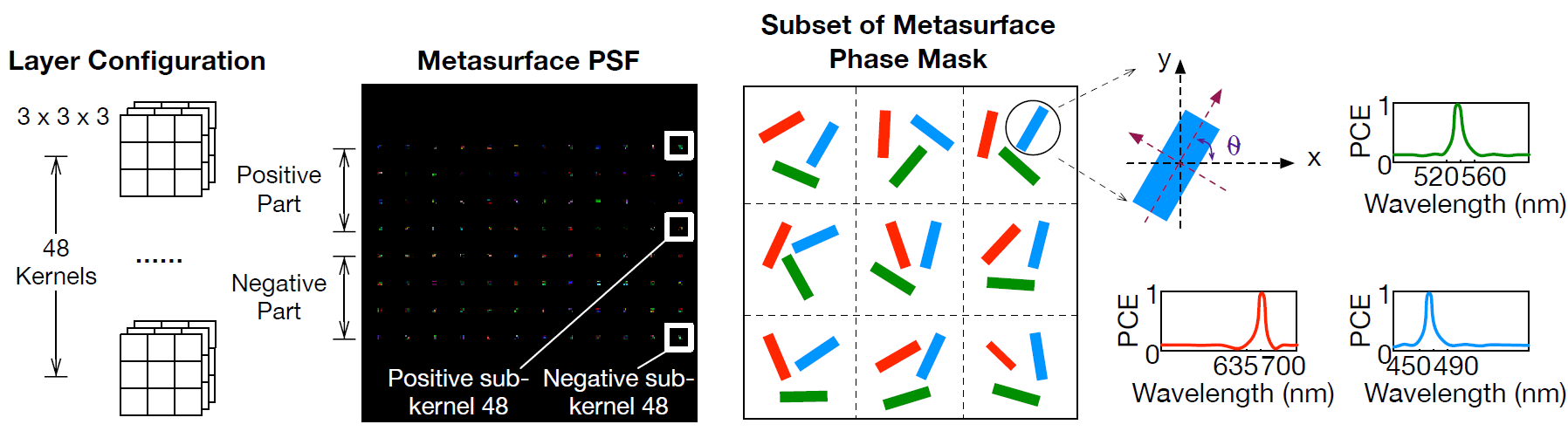}
  \caption{The target PSF for a convolutional layer that has 48 kernels, where each kernel has 3 channels and each channel has 3$\times$3 elements. The PSF is arranged as 12$\times$8 tiles, with 1 positive and 1 negative tile for each kernel. Each tile is 3$\times$3. Each tile has 3 channels, corresponding to different frequencies of incident light. On the metasurface, each meta-element has a unique in-plane rotation $\theta$. Each unit cell in the metasurface contains 3 kinds of meta-elements with distinct geometric parameters, and each kind is associated to a different channel. Following Eq. (\ref{eq:psf}), all the meta-elements associated to the same channel collectively form a phase mask that shapes the PSF for that channel, which matches the corresponding channel of the convolution kernels. Meta-elements associated to the same channel have the same PCE spectrum, and in the ideal case, the corresponding PSF is applied to the associated input channel only, without cross-talk.}
  \label{fig:ms}
\end{figure*}



\subsection{Single-channel case}
\label{sec:sys:ms:phase}

Our metasurface phase mask consists of a set of meta-elements, each of which induces a particular phase change to the incident light. Physically, we use the Pancharatnam-Berry (P-B) phase~\cite{meta-review, MetaPol}. P-B phase comes from the spin rotation effect that arises from the interaction between circularly-polarized light and an anisotropic meta-element~\cite{meta-hol-review}.
Specifically, a meta-element that is geometrically symmetric with respect to its main axes and that has an in-plane orientation angle $\theta$ induces a geometric phase $2\theta$ on the incident light~\cite{meta-achromat, meta-hol-review}. Therefore, for a phase mask that consists of $M\times N$ meta-elements, each having an in-plane rotation of $\theta_{m, n}$, the phase profile $\Phi$ of the phase mask is also a $M\times N$ matrix, where each element is $2\times \theta_{m, n}$.

In a 4-$f$ system (with a DOE or a metasurface), the PSF relates to the phase profile ($\Phi$) of the phase mask according to the following equation:

\begin{equation}
  \label{eq:psf}
  \mathbf{PSF}(\Phi) = \left\vert\mathcal{F}^{-1}\left\lbrace e^{i\Phi} \right\rbrace  \right\vert^{2},
\end{equation}

\noindent where $\mathcal{F}^{-1}$ denotes the inverse 2D Fourier transform, and $e^{i\Phi}$ acts as the system's Transfer Function\cite{goodman2005introduction}. Thus, finding the optimal phase profile is formulated as the following optimization problem:

\begin{equation}
  \label{eq:phase_opt}
  \underset{\Phi_{k}}{\mathrm{min}}\: \vert\vert\, W_{k} - \vert\mathcal{F}^{-1}\left\lbrace e^{i\Phi_{k}} \right\rbrace  \vert^{2}\,  \vert\vert_{F}^{2}\,,
\end{equation}

\noindent where $\vert\vert\, \cdot \,\vert\vert_{F}$ denotes the Frobenius norm, $W_{k}$ denotes the $k^{th}$ channel of the target PSF that encodes the convolution kernels, and $\Phi_{k}$ denotes the discretized 2D phase profile of channel $k$.

%
%


When performing the phase optimization process, we use a discretized version of the target PSF, encoding it in a numerical array of $N\times N$ pixels. During this phase optimization process, we iteratively update the numerical array $\Phi_{k}$ of $N\times N$ pixels that contains the phase profile. The optimization process has the objective of making the PSF yielded by $\Phi_{k}$ be as close as possible to the target PSF that encodes $W_{k}$. Once we obtain the optimized numerical array $\Phi_{k}$, we can do the mapping between the (row, column) positions in the numerical phase array and $(u,\,v)$ positions in the plane of the phase mask.
For this, we remember that each point $(u,\,v)$ in the Fourier's plane  (where the phase mask lies) is associated with a point $(f_{\eta},\,f_{\xi})$ in the frequency space domain given by $(f_{\eta},\,f_{\xi}) = \left(\frac{u}{\lambda f},\,\frac{v}{\lambda f}\right)$, as can be seen in \Equ{eq:FourierPlane}. Additionally, we know that the largest (absolute value of) spatial frequency component of the input image that makes it into the Fourier plane is given by one half of the sampling rate of the display in the input plane, known as the Nyquist rate \cite{goodman2005introduction}. 
That is:
\begin{equation}
\label{eq:MaxFreq}
\left\vert f_{\eta} \right\vert\sbl{max} = \left\vert f_{\xi} \right\vert\sbl{max} = \frac{1}{2d},
\end{equation}
where $d$ is the pixel pitch of the display in the input plane that projects the input image into the optical system.
Given this, the size of the spatial frequency domain is $1/d$, and since the phase mask needs to cover it all, the size of the phase mask needs to be $\frac{\lambda f}{d}$.
We then divide the region of $\frac{\lambda f}{d}\times\frac{\lambda f}{d}$ into $N\times N$ subregions (or ``pixels''). 
Thus, the size of these ``phase pixels'' is given by:
\begin{equation}
\label{eq:PhasePixelSize}
\Delta u = \frac{\lambda f}{Nd}.
\end{equation}

After we have obtained the numerical values of the elements in the optimized $\Phi_{k}$ array, we simply divide them by 2 to obtain the value of the in-plane orientation of each meta-element on the metasurface. We would like to note that the objective function in \Equ{eq:phase_opt} is differentiable. Thus, phases (and therefore, the values of the meta-elements' orientations) could be co-optimized with the rest of DNN in an end-to-end fashion, as we will show in \Sect{sec:sys:fm}.


Now, we'll discuss how we design the target PSF that encodes the network's convolution kernels. Since PSF physically corresponds to light intensity~\cite{goodman2005introduction} and there is no negative light intensity, the negative values in a kernel cannot be physically implemented. A mathematical explanation is that the PSF is the square form of an expression, as shown in \Equ{eq:psf}, and thus must be non-negative. To address this limitation, we use the idea of Chang et. al.~\cite{chang2018hybrid} by splitting a kernel into two \textit{sub-kernels}; one contains only the positive portion and the other contains only the (absolute value of the) negative portion of the kernel. The final kernel is calculated by subtracting the negative sub-kernel from the positive sub-kernel.

As a result, for a layer that has $K$ kernels, the target PSF contains $2K$ tiles that are used to represent these kernels. \Fig{fig:ms} shows the target 3-channel PSF for a layer that has 48 kernels, each of dimension 3$\times$3$\times$3. The PSF has 96 tiles, where the ones in the top half are the positive sub-kernels, and the ones in the bottom half contain the are sub-kernels. Each tile is 3$\times$3, which corresponds to the 3$\times$3 elements in one channel of a 3-channel sub-kernel. In order to yield this target PSF, the in-plane rotation of every meta-element on the metasurface is optimized from \Equ{eq:phase_opt}. The three meta-elements in a unit cell of the metasurface are used to yield the three channels of the target multi-channel PSF, which we discuss next.

\subsection{Multi-channel case}
\label{sec:sys:ms:pol}

With what was discussed above, we are able to use the in-plane rotations of the meta-elements to engineer the PSF to match any given 2D matrix (i.e., a channel in the target kernel). When going from the single-channel case to the multi-channel case, we first note that the size of the ``phase pixels'' will depend on the wavelength $\lambda$ according to \Equ{eq:PhasePixelSize}. This means that the pixels (regions in the metasurface that impart a given phase value to incident light) associated with different channels will have different sizes. We address this by noting that for our choice for the values of $f$, $d$, and $\lambda$ range, the size of the metasurface phase pixels for all channels will be larger than the size of the metasurface's unit cells (the former being in the order of microns and the latter being in the order of hundreds of nanometers). This means that the phase pixels of any given channel are filled with multiple unit cells (each of which contains one of each kind of meta-element associated to the different channels). Since the orientation of each meta-element in the unit cell is independent to that of the others, we can essentially multiplex the phase profiles associated to the different channels on the same metasurface without issues.

The remaining challenge is making sure that the PSF that is yielded by the set of meta-elements of one kind should only be applied to light of the corresponding channel (frequency component) in order to correctly implement the convolution layer. That is, we must minimize cross-talk effects that make meta-elements apply the geometrical phase to a portion of the light of the other channels.
We make use of the fact that each meta-element acts as a phase retarder, which makes a portion of incident circularly-polarized light to change its handedness and get embued with a geometrical phase. This portion is dependent on the wavelength of incident light, and is the
polarization conversion efficiency (PCE). Effectively, when incident light from the input image interacts with the set of meta-elements that has a given PCE spectrum, the corresponding intensity profile in the output plane (given by the convolution between the input intensity profile and the PSF yielded by this set of meta-elements)  is weighted by $\mathrm{PCE}(\omega)$ at frequency $\omega$. With the PCE taken into account, the contribution of intensity in the output (capture) plane that is attributed to an input channel $c$ can be expressed as:

\begin{equation}
\label{eq:channel_contrib}
  I_{con}^{c}(x,\,y,\omega) = \sum_{c'=1}^{C_{in}}\mathbf{PCE}^{c'}(\omega)(\mathbf{PSF}^{c'}(\eta,\,\xi)\star I_{in}^{c}(\eta,\,\xi,\,\omega)),
\end{equation}

\noindent where channel $c$ of the input image is affected by the PSFs $c'$ yielded by the different meta-element sets in the metasurface, which is where the sum over PSF channels $c'$ comes from.

In the ideal case, the PCE is a delta function that is 1 for a very narrow band, e.g., red wavelength, and 0 for all other wavelengths. That way, the PSF carried by the meta-elements will effectively convolve only with one particular input channel. \Fig{fig:ms} shows an example that supports 3 input channels (e.g., natural light from the real-world). The key is that each unit cell consists of three meta-elements, each carrying a unique PCE. When white light containing the RGB wavelengths (channels) is incident on the phase mask, the red channel will acquire the phase imparted by the meta-element associated to the red-channel only, mimicking the convolution between the red channel of the input and the corresponding PSF; the same process applies to green and blue wavelengths as well. The output is naturally the superposition (sum of intensity) of the three convolution outputs.

In reality, the metasurface design is regulated by fabrication constraints and, thus, the actual PCE will never be an ideal delta function, which leads to cross-talk, i.e., the PSF of a channel $i$ will be convolved with input channel $j$. Therefore, it is critical to optimize the end-to-end system to minimize the impact of cross-talk, which we discuss next.

\section{End-to-End Training Framework}
\label{sec:sys:fm}

\begin{figure*}[t]
  \centering
  \captionsetup{width=\columnwidth}
  \includegraphics[width=0.96\columnwidth]{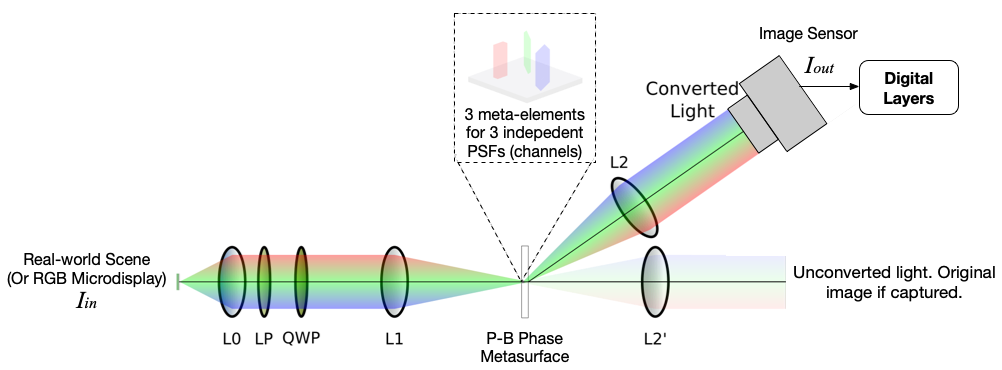}
  \caption{Hardware system. Light from the real-world (or a microdisplay) is collimated by lens L0, and then goes through a linear polarizer (LP) and a broadband quarter-wave plate (QWP) so that light becomes circularly-polarized. The plane of the LP is the input plane of the 4$f$ system. A metasurface phase mask is placed at the Fourier plane. A portion of the light gets the engineered phase shift and has its polarization handedness switched. This ``converted'' portion of the light gets sent to lens L2, via a phase ``grating''~\cite{palmer2002diffraction} added to the phase profile, while the unconverted light is sent to lens L2' on the optical axis. L2 is at a distance $f$ from the metasurface. Finally, light is detected by the image sensor placed at a distance $f$ after L2.}
\label{fig:hw}
\end{figure*}

\begin{figure*}[t]
\centering
\captionsetup{width=\columnwidth}
\includegraphics[width=1.0\columnwidth]{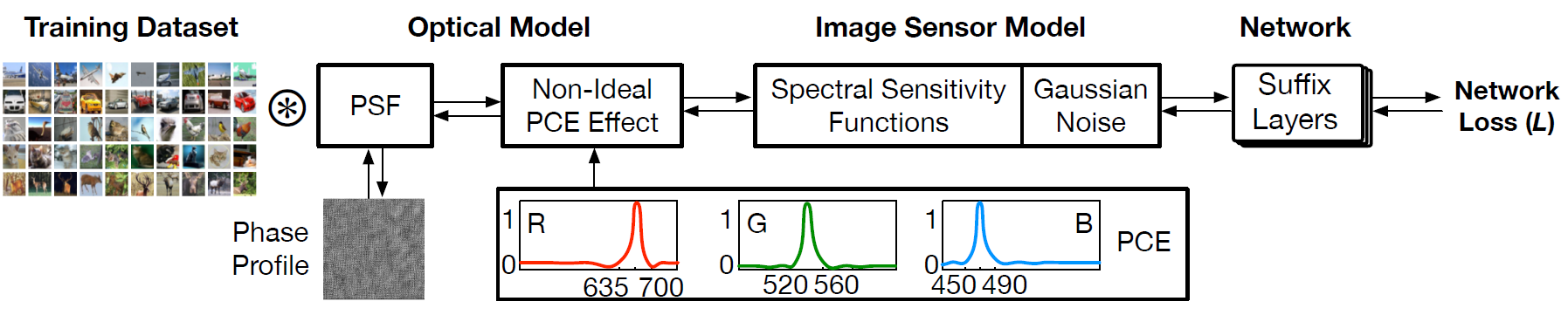}
\caption{End-to-end training framework. The polarization conversion efficiency profile is network-independent, and is obtained from real metasurface optics fabrications. The phase profile can be optimized using the training data and fine-tuned with the DNN model suffix layers (i.e., layers implemented in the digital domain) together in an end-to-end fashion.}
\label{fig:e2e}
\end{figure*}

\Fig{fig:hw} shows the end-to-end system setup, from light in the real-world scene (or equivalently a RGB microdisplay) to the network output. It consists primarily of three components: a metasurface-based optical system that implements the first convolution layer of a DNN, an image sensor that captures the output of the optical convolution, and a digital processor that finishes the rest of the network, which we call the ``suffix layers.''

In order to maximize the overall network accuracy, we co-optimize the optical convolution layer with digital layers in an end-to-end fashion. \Fig{fig:e2e} shows the end-to-end differentiable pipeline, which incorporates three components: an optical model, a sensor model, and the digital network. The phase profile ($\Phi$) in the optical model and the suffix layer weights ($W$) are optimization variables. The loss function $L$ that is used to measure this system's performance is the same as what can be used with a conventional all-digital neural network, like the log loss function.

The input to the system is light, which in practical applications are from the real-world scene. In our work, we envision a RGB microdisplay displaying images from datasets to emulate the real-world scene as is done in prior work~\cite{chang2018hybrid}. Thus, channel $c$ in the input is expressed as:

\begin{equation}
\label{eq:input_im}
I_{in}^{c}(x,\,y,\,\omega) = \mathbf{SPD}^{c}(\omega) I_{in}^{c}(x,\,y),
\end{equation}

\noindent where SPD$^{c}(\omega)$ is the spectral power distribution of color channel $c$ in the input light at frequency $\omega$, and $I_{in}^{c}(x,\,y)$ is the light intensity of pixel ($x$, $y$) in channel $c$ and is directly read from the RGB image in the training data.

In the forward pass during training (and also inference), the PSF generated from the phase profile ($\Phi$) is first convolved with the input image from a training dataset under the impact of the PCE as described by \Equ{eq:channel_contrib}. The convolution output is captured by the image sensor, which we model by its spectral sensitivity function (SSF). In addition, image sensors introduce different forms of noise, which we collectively model as a Gaussian noise function $N(\mu, \sigma^2)$, based on the Central Limit Theorem. In order to retain the behavior of optical shot noise, which is proportional to the number of incident photons, we programmed the noise layers on our DNN to have a standard deviation $\sigma$ that is proportional to the magnitude of those layers' inputs. 

As a result, the output signal yielded by the sensor is given by:

\begin{equation}
\label{eq:total_output}
I_{out}(x,\,y) = \sum_{c=1}^{C_{in}}\int \mathbf{SSF}(\omega) I_{con}^{c}(x,\,y,\,\omega)d\omega + N(\mu, \sigma^2).
\end{equation}

Note that the sum across different input channels, required by correctly implementing a convolution layer, is realized in our design ``for free'' by superimposing the intensity profile of the different frequency components that are incident on the detector. The total light intensity of at the output plane is naturally the (weighted) sum of the intensities of individual channel components.

Plugging \Equ{eq:input_im} into \Equ{eq:channel_contrib} and plugging \Equ{eq:channel_contrib} into \Equ{eq:total_output} yields the sensor output, which becomes the input to the digital suffix layers to then produce the network's output and loss. The sensor output can then be expressed as:
\begin{equation}
\label{eq:total_output_crosstalk}
I_{out}(x,\,y) = \sum_{c}^{C_{in}}\left[\sum_{c'}^{C_{in}}\left(\int \mathbf{SSF}(\omega)\mathbf{PCE}^{c'}(\omega)\mathbf{SPD}^{c}(\omega)d\omega\right)\mathbf{PSF}^{c'}(\eta,\xi)\right]\star I_{in}^{c}(\eta,\,\xi) + N(\mu, \sigma^2).
\end{equation}

The above expression can be simplified into the form:
\begin{equation}
\label{eq:total_output_crosstalk_simp}
I_{out}(x,\,y) = \sum_{c}^{C_{in}}\left[\sum_{c'}^{C_{in}}A_{c'c}\mathbf{PSF}^{c'}(\eta,\xi)\right]\star I_{in}^{c}(\eta,\,\xi) + N(\mu, \sigma^2),
\end{equation}

\noindent where $A_{c'c}=\int \mathbf{SSF}(\omega)\mathbf{PCE}^{c'}(\omega)\mathbf{SPD}^{c}(\omega)d\omega$, and we can define a \textit{cross-talk matrix} whose elements are $A_{c'c}$ for the different values of $c'$ and $c$. Note that the convolution layer mapped to metasurface, as described by \Equ{eq:total_output}, is differentiable, and thus enables end-to-end training using classic gradient descent-based optimization methods. In doing so, the suffix layers can be optimized to compensate for the effects introduced by cross-talk. During back-propagation, the phase profile ($\Phi$) and the digital layer weights ($W$) are optimized. Note that the PCEs of the meta-elements are network-independent, and are directly obtained from fabrication. They are kept fixed throughout training.

\paragraph{Practical Considerations} Since different sub-kernels are tiled together for the ease of training, the outputs of different sub-convolutions (i.e., convolution between a sub-kernel and the corresponding channel in the input) will be tiled too. We must make sure that the different sub-convolution outputs do not overlap so that we can assemble them (digitally) to form the actual convolution output. To that end, we pad each sub-kernel with space between cells in the phase mask. For instance in the illustrative example shown in \Fig{fig:ms}, we pad each 3$\times$3 sub-kernel by 8 pixels each side to form a 19$\times$19 matrix, effectively separating neighboring sub-kernels in the phase mask by 16 pixels. In our work, we pad the sub-kernels by 20 pixels each side to form a 43$\times$43 matrix, effectively neighboring sub-kernels by 40 pixels. In summary, given the kernels in a target convolution layer, we first separate each kernel into two sub-kernels, tile all the sub-kernels on the same plane, and then add padding to form the actual target PSF matrix, for which we then apply the phase optimization described in \Equ{eq:phase_opt} to derive the required phase profile, and thus the in-plane rotation of each meta-element on the metasurface.

\paragraph{Monochromatic Sensor} It is important to note that our system design lets us use a simple monochromatic sensor rather than a conventional RGB sensor. This is because of two reasons. First, the results of all output channels lie completely within a 2D plane, because the tiles of different convolution kernels and of different channels within a kernel are tiled in a 2D plane (\Fig{fig:ms}). Second, since the monochromatic sensor integrates the intensity of all incident frequency components, this naturally emulates the sum over channels that needs to be performed when calculating the output of a digital convoulutional layer, as described by \Equ{eq:cnn}. 

\section{Evaluation}
\label{sec:eval}

\subsection{Methodology}
\label{sec:eval:method}

\paragraph{Networks and Datasets} We implement the end-to-end training framework in TensorFlow. We differentiate the metasurface optics layer (from phase profile to the target kernel) and integrate the differentiation into the back-propagation process. We evaluate on three networks: AlexNet~\cite{alexnet}, VGG-16~\cite{simonyan2014very}, and ResNet50~\cite{resnet}, all on the CIFAR-100 dataset~\cite{cifar}.

\begin{figure}[t]
    \centering
    \includegraphics[width=.96\columnwidth]{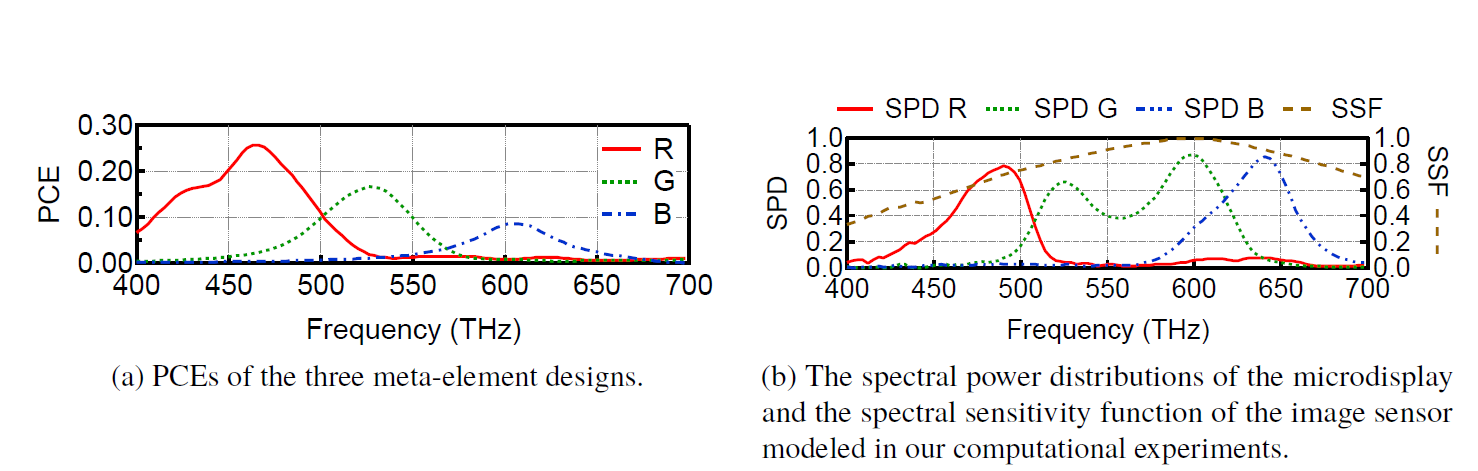}
    \captionsetup{width=\linewidth}
    \caption{Parameters of hardware components. We use these to compute the cross-talk matrix, which is necessary to account for the effects of the cross-talk between channels.}
    \label{fig:para}
\end{figure}

\paragraph{Training Details} To accelerate convergence, we first train the entire network in three separate steps: 1) train the entire network digitally, 2) use the first layer's weights to optimize the phase mask, and 3) finally the optimized phase mask and the pre-trained suffix layers are then fine-tuned end-to-end until convergence. The phase optimization process consists on running a gradient descent algorithm to minimize the loss function shown in \Equ{eq:phase_opt}. In the Supplementary Information, we provide further details about the derivation of the expression of the gradient of this function, as well as a way for it to be computed efficiently. In all networks, the results converge after 2,000 iterations of the gradient descent cycle. The phase masks and suffix layers' parameters are then end-to-end fine-tuned with the suffix layers for 50, 150, and 50 epochs for AlexNet, VGG, and ResNet, respectively.

\paragraph{Hardware Modeling Details} The metasurface designs we use are from Wang et al.~\cite{wang2016visible}, which provide the detailed design specifications of three kinds of meta-elements, each of which is designed to have their highest response at red, green, and blue bands of the visible spectrum. The meta-elements are pillars made out of $\mathrm{Si}$, with a height of $320\,\mathrm{nm}$, sitting on a $\mathrm{SiO_{2}}$ substrate. These meta-elements have a rectangular cross-section whose dimensions depend on the channel each meta-element is associated with. Meta-elements associated with the red channel have a cross section of $145\,\mathrm{nm}\times105\,\mathrm{nm}$, those associated with the green channel have a cross section of $110\,\mathrm{nm}\times80\,\mathrm{nm}$, and associated with the blue channel have a cross section of $90\,\mathrm{nm}\times45\,\mathrm{nm}$. A unit cell of the metasurface is $420\,\mathrm{nm}\times420\,\mathrm{nm}$ and contains the three kinds of meta-elements on it. Based on these design specifications, we run the well-known MEEP package~\cite{oskooi2010meep, meep} for electromagnetics simulations to generate the corresponding PCEs of the three kinds of meta-elements, which we use in our design and simulation. \Fig{fig:para}a shows the PCE of the three kinds of meta-elements. While the PCEs are not ideal delta functions, they do provide the highest responses in distinct bands.

We use the Sony ICX274AL image sensor, whose SSF we adapt from the design specification~\cite{sonyICX274AL}. We use the eMagin OLED-XL RGB microdisplay from eMagin, whose SPDs we obtained by contacting the display manufacturer. \Fig{fig:para}b shows the normalized SPDs of the microdisplay pixels (left $y$-axis) and the SSF of the sensor (right $y$-axis). Comparing the SPDs and the PCEs in \Fig{fig:para}, the PCE and SPD peaks are close to each other (e.g., both the red PCE and the red light from the microdisplay peak at around 450 THz to 500 THz). We use these PCE and SPD functions to calculate the cross-talk matrix, which is a necessary part in the system's optimization process so that it can account for cross-talk effects.

\begin{figure}[t]
  \centering
  \includegraphics[width=.7\columnwidth]{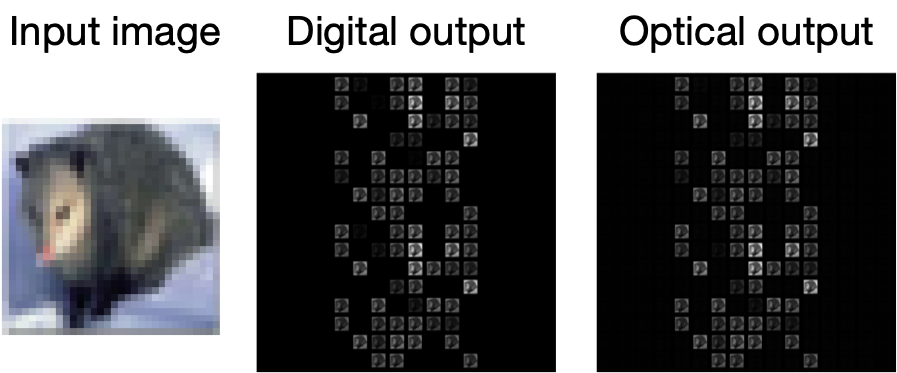}
  \caption{Comparison between the output of a convolution layer when implemented digitally, and when implemented optically. The optical output is simulated, taking the effects of cross-talk between channels into account}
  \label{fig:output}
\end{figure}

\begin{figure}[t]
    \centering
    \includegraphics[width=.96\columnwidth]{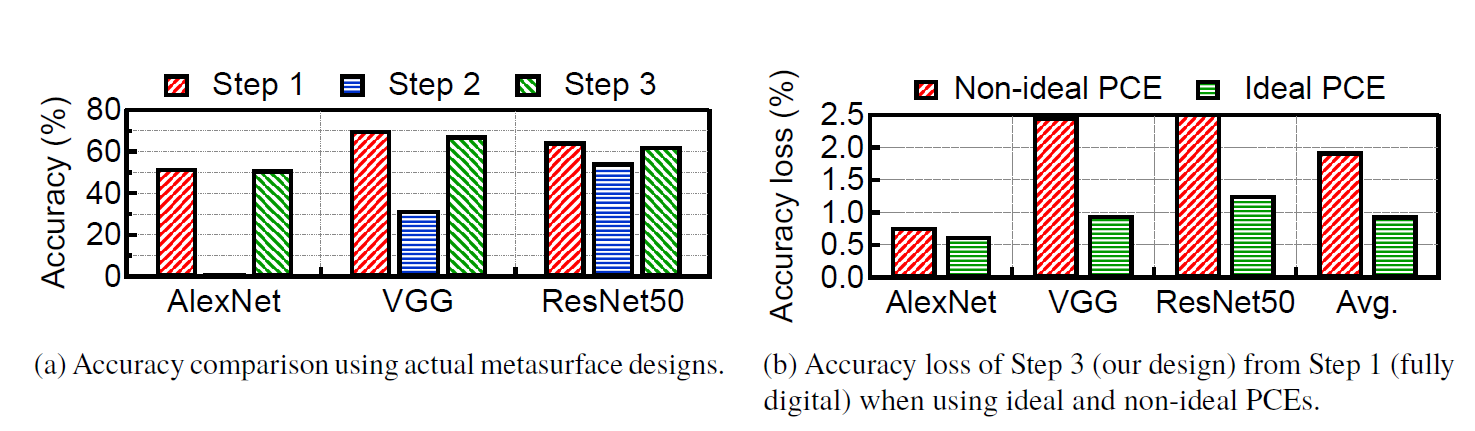}
    \captionsetup{width=\linewidth}
    \caption{Accuracy results obtained after running the pipeline on different datasets.}
    \label{fig:results}
\end{figure}

\subsection{Network Accuracy and Efficiency}

\paragraph{Phase Optimization} We find that our phase optimization allows us to generate a PSF that resembles the target kernel. The loss decreases by two to three orders of magnitude from its initial value after 2,000 iterations, reaching convergence. \Fig{fig:output} compares the output feature map of the first layer between the digital implementation and the optical implementation, which match well, indicating that the optical output provides a desirable input to the suffix digital layers. Note that the different output channels are tiled in one 2D plane.

\paragraph{Accuracy} We find that our end-to-end design leads to little accuracy loss. \Fig{fig:results}a shows the network accuracies of the three training steps. As we can see, step 1 is the original network executed fully in the digital domain, and generally has the highest accuracy. Step 2 optimizes the PSF from the target convolution kernel and directly integrates the optimized PSF into the network; this leads to significant lower accuracy. However, after end-to-end fine-tuning with the suffix layers, we could largely gain the accuracy loss. Overall across the three networks, our optical-electrical DNN system leads to on average only an 1.9\% (0.8\% minimum) accuracy loss compared to fully-digital networks. The accuracy loss could be further improved if we could improve the PCE of the metasurface designs. \Fig{fig:results}b compares the accuracy loss on the three networks between using ideal PCEs (i.e., delta functions) and non-ideal PCEs (from actual designs, used in \Fig{fig:results}a). Using an ideal PCE leads to an average accuracy loss of only 0.9\%. This suggests that our system design can readily benefit from better metasurface designs in the future.

\paragraph{Efficiency} Our system reduces the overall computational cost in two ways. First, we completely avoid the need to process raw sensor data using the power-hungry ISP. Traditional vision systems use ISP to process raw sensor data (e.g., demosaic, tone mapping) to generate RGB images for the DNN to process, but since in our system the image sensor captures the output feature map rather than the RGB image, the ISP processing could be avoided.

We measure the power of the Nvidia's Jetson TX2 board, which is representative of today's high-end systems for mobile vision~\cite{tx2spec}. The ISP consumes about 0.5 mJ of energy to generate one RGB frame. In contrast, the energy consumed to execute AlexNet, VGG, and ResNet50 on a Google TPU-like specialized DNN processor~\cite{jouppi2017datacenter} is 0.31 mJ, 0.36 mJ, 0.05 mJ, respectively, which we estimate using a recent DNN energy modeling framework~\cite{feng2019asv, asvgithub}. Avoiding the ISP yields 2.4$\times$ to one order of magnitude energy saving.

Secondarily, by pushing the first CNN layer into the  optical domain, we also reduce the total amount of computations (measured in the number of multiply-accumulate, or MAC, operations) to compute the DNN. In AlexNet, the MAC saving is 4.4\%, and the MAC saving is 0.8\% and 0.5\% for ResNet50 and VGG, respectively.

\section{Discussion and Conclusion}
\label{sec:disc}

This paper demonstrates the first free-space optics approach that enables general convolutional layers. The significance our system is three-fold. First, it eliminates the need for the power-hungry ISP. Second, it simplifies the sensor design and enables the use of a simple monochromatic sensor rather than a RGB sensor. Finally, it reduces the computation to execute a DNN. Overall, we show an order of magnitude of energy saving with 1.9\% on average (0.8\% minimum) accuracy.

We are in the process of fabricating an optical prototype. While our current design pushes only the first DNN layer into optics, pushing more layers into optics is possible but not without challenges. The key challenge is that it is difficult to optically implement non-linear functions, which usually have to be implemented in the electrical domain, introducing optical-electrical interface transduction overhead. Therefore, a future direction is to comprehensively explore the hybrid optical-electrical design space to evaluate the optimal partition between optics and electronics.

\section{Funding, Acknowledgments, and Disclosures}
\label{sec:ack}
\textbf{Funding.} University of Rochester (UR) University Research Award.\\
\textbf{Acknowledgments.} C. Villegas Burgos would like to thank the Science and Technology Council of Mexico (Consejo Nacional de Ciencia y Tecnología, CONACYT) for the financial support provided by the fellowship they granted him. \\
\textbf{Disclosures.} The authors declare no conflict of interest. 

\section*{Supplementary Materials}
\label{sec:supp}
See Supplement 1 for supporting content.

\bibliography{refs}


\section*{Supplementary material}

# A Design Framework for Metasurface Optics-based Convolutional Neural Networks: Supplemental Document

In this Supplemental Document, we show in detail the derivations for the equations we use in the gradient descent algorithm that performs the phase optimization process described in the main text. Additionally, we also derive the expressions for the gradients we need to compute when performing the end-to-end fine-tuning process where the phase profile is trained at the same time as the suffix layers of the hybrid neural network.

## 1. PHASE OPTIMIZATION PROCESS

In the phase optimization process, we use a gradient descent method to iteratively update the phase profile $\Phi$ in order to minimize a loss function $L_\Phi$ given by:

$$L_\Phi(\Phi; \mathbf{PSF}_{\text{target}}) = \left|\left| \mathbf{PSF}_{\text{target}} - \left|\mathcal{F}^{-1}\left\{\exp\left(i\Phi\right)\right\}\right|^2 \right|\right|^2, \tag{S1}$$

where $||\cdot||^2$ denotes the Frobenius norm of the enclosed numerical array.

For simplicity in our derivation, we use $\mathbf{PSF}_{\text{target}} = Y$ onward. Here, both $Y$ and $\Phi$ are three-dimensional numerical arrays, where the first two dimensions correspond to spatial positions and the third dimension corresponds to channels. By using the definition of the Frobenius norm and the (discrete) inverse Fourier transform, we have:

$$L_\Phi(\Phi; Y) = \sum_{k,l,c} \left( Y^c_{kl} - \left| \sum_{m}^{M-1} \sum_{n}^{N-1} \exp\left(i\Phi^c_{mn}\right) \exp\left(-2\pi i \left(\frac{mk}{M} + \frac{nl}{N}\right)\right) \right|^2 \right)^2, \tag{S2}$$

where the $c$ superscripts denote the channel index, and $M$ and $N$ are the number of rows and columns in the phase profile numerical array.

In our phase optimization process, we minimize function $L_\Phi$ via gradient descent. This means we iteratively calculate the gradient of $L_\Phi$ with respect to $\Phi$, denoted $\nabla_\Phi L_\Phi$, and update the values in $\Phi$ as $\Phi = \Phi - \alpha \nabla_\Phi L_\Phi$ with $\alpha$ being a "learning rate". We derived an expression for $\nabla_\Phi L_\Phi$ that allows us to compute this gradient in terms of operations between numerical arrays which can be efficiently carried out by libraries such as NumPy or Tensorflow. We will derive it below.

We calculate the partial derivative of $L_\Phi$ with respect to an element in the phase profile array, $\Phi^c_{mn}$. After a few straight-forward simplifications, we find:

$$\frac{\partial L_\Phi}{\partial \Phi^c_{mn}} = -2 \sum_{k,l} \left[ \left( Y^c_{kl} - \left|\mathcal{F}^{-1}\left\{\exp\left(i\Phi^c\right)\right\}_{kl}\right|^2 \right) \frac{\partial}{\partial \Phi^c_{mn}} \left| \sum_{m'}^{M-1} \sum_{n'}^{N-1} \exp\left(i\Phi^c_{m'n'}\right) \exp\left(-2\pi i \left(\frac{m'k}{M} + \frac{n'l}{N}\right)\right) \right|^2 \right]$$

$$\frac{\partial L_\Phi}{\partial \Phi^c_{mn}} = -2 \sum_{k,l} \left[ \left( Y^c_{kl} - \left|\mathcal{F}^{-1}\left\{\exp\left(i\Phi^c\right)\right\}_{kl}\right|^2 \right) \frac{\partial}{\partial \Phi^c_{mn}} \left|\mathcal{F}^{-1}\left\{\exp\left(i\Phi^c\right)\right\}_{kl}\right|^2 \right]$$

$$\tag{S3}$$

Before we continue further, we first calculate separately the result of the partial derivative that appears on the right side of Eq. (S3).

We have:

$$\frac{\partial}{\partial \Phi^c_{mn}} \left| \mathcal{F}^{-1}\{\exp(i\Phi^c)\}_{kl} \right|^2 = \frac{\partial}{\partial \Phi^c_{mn}} \left( \text{Re}\left[\mathcal{F}^{-1}\{\exp(i\Phi^c)\}_{kl}\right] \right)^2 + \frac{\partial}{\partial \Phi^c_{mn}} \left( \text{Im}\left[\mathcal{F}^{-1}\{\exp(i\Phi^c)\}_{kl}\right] \right)^2$$

$$\frac{\partial}{\partial \Phi^c_{mn}} \left| \mathcal{F}^{-1}\{\exp(i\Phi^c)\}_{kl} \right|^2 = \frac{\partial}{\partial \Phi^c_{mn}} \left( \sum_{m'}^{M-1} \sum_{n'}^{N-1} \cos\left( \Phi^c_{m'n'} - 2\pi\left(\frac{m'k}{M} + \frac{n'l}{N}\right) \right) \right)^2$$

$$+ \frac{\partial}{\partial \Phi^c_{mn}} \left( \sum_{m'}^{M-1} \sum_{n'}^{N-1} \sin\left( \Phi^c_{m'n'} - 2\pi\left(\frac{m'k}{M} + \frac{n'l}{N}\right) \right) \right)^2$$

$$\frac{\partial}{\partial \Phi^c_{mn}} \left| \mathcal{F}^{-1}\{\exp(i\Phi^c)\}_{kl} \right|^2 = -2\text{Re}\left[\mathcal{F}^{-1}\{\exp(i\Phi^c)\}_{kl}\right] \sin\left( \Phi^c_{mn} - 2\pi\left(\frac{mk}{M} + \frac{nl}{N}\right) \right)$$

$$+ 2\text{Im}\left[\mathcal{F}^{-1}\{\exp(i\Phi^c)\}_{kl}\right] \cos\left( \Phi^c_{mn} - 2\pi\left(\frac{mk}{M} + \frac{nl}{N}\right) \right)$$

$$\frac{\partial}{\partial \Phi^c_{mn}} \left| \mathcal{F}^{-1}\{\exp(i\Phi^c)\}_{kl} \right|^2 = 2\text{Re}\left[\mathcal{F}^{-1}\{\exp(i\Phi^c)\}_{kl}\right] \text{Im}\left[\exp\left(-i\Phi^c_{mn} + 2\pi i\left(\frac{mk}{M} + \frac{nl}{N}\right)\right)\right]$$

$$+ 2\text{Im}\left[\mathcal{F}^{-1}\{\exp(i\Phi^c)\}_{kl}\right] \text{Re}\left[\exp\left(-i\Phi^c_{mn} + 2\pi i\left(\frac{mk}{M} + \frac{nl}{N}\right)\right)\right]$$

$$\frac{\partial}{\partial \Phi^c_{mn}} \left| \mathcal{F}^{-1}\{\exp(i\Phi^c)\}_{kl} \right|^2 = 2\text{Im}\left[\mathcal{F}^{-1}\{\exp(i\Phi^c)\}_{kl} \exp\left(-i\Phi^c_{mn} + 2\pi i\left(\frac{mk}{M} + \frac{nl}{N}\right)\right)\right] \quad (S4)$$

Substituting Eq. (S4) into Eq. (S3), we have:

$$\frac{\partial L_\Phi}{\partial \Phi^c_{mn}} = -4 \sum_{k,l} \left( Y^c_{kl} - \left|\mathcal{F}^{-1}\{\exp(i\Phi^c)\}_{kl}\right|^2 \right) \text{Im}\left[\mathcal{F}^{-1}\{\exp(i\Phi^c)\}_{kl} \exp\left(-i\Phi^c_{mn} + 2\pi i\left(\frac{mk}{M} + \frac{nl}{N}\right)\right)\right]$$

$$\frac{\partial L_\Phi}{\partial \Phi^c_{mn}} = -4\text{Im}\left[ \exp(-i\Phi^c_{mn}) \sum_{k,l} \left( Y^c_{kl} - \left|\mathcal{F}^{-1}\{\exp(i\Phi^c)\}_{kl}\right|^2 \right) \mathcal{F}^{-1}\{\exp(i\Phi^c)\}_{kl} \exp\left(+2\pi i\left(\frac{mk}{M} + \frac{nl}{N}\right)\right) \right]$$

$$\frac{\partial L_\Phi}{\partial \Phi^c_{mn}} = -4\text{Im}\left[ \exp(-i\Phi^c_{mn}) \mathcal{F}_{2D}\left\{ \left( Y - \left|\mathcal{F}^{-1}_{2D}\{\exp(i\Phi)\}\right|^2 \right) \mathcal{F}^{-1}_{2D}\{\exp(i\Phi)\} \right\}_{mnc} \right]. \quad (S5)$$

With Eq. (S5), we can finally express the gradient of $L_\Phi$ with respect to $\Phi$ as:

$$\nabla_\Phi L_\Phi = -4\text{Im}\left[ e^{-i\Phi} \mathcal{F}_{2D}\left\{ \left( Y - \left|\mathcal{F}^{-1}_{2D}\{e^{i\Phi}\}\right|^2 \right) \mathcal{F}^{-1}_{2D}\{e^{i\Phi}\} \right\} \right]. \quad (S6)$$

This is the expression we use to efficiently compute $\nabla_\Phi L_\Phi$ during the gradient descent optimization process to find the phase profile $\Phi$ whose yielded PSF approaches the target PSF that encodes the convolutional kernel tensor of the replaced convolutional layer.

## 2. EXPRESSIONS FOR THE NEURAL NETWORK'S GRADIENTS

Recall from the main text that the output signal yielded by the photodetector in our system is given by:

$$I_{out}(x, y) = \sum_c^{C_{in}} \left[ \sum_{c'}^{C_{in}} \left( \int \mathbf{SSF}(\omega) \mathbf{PCE}^{c'}(\omega) \mathbf{SPD}^c(\omega) d\omega \right) \mathbf{PSF}^{c'}(\eta, \xi) \right] \star I^c_{in}(\eta, \xi), \quad (S7)$$

where $\mathbf{SSF}(\omega)$ is the photodetector's spectral sensitivity function, $\mathbf{PCE}^{c'}(\omega)$ is the polarization conversion efficiency spectrum of the metasurface's meta-elements associated with channel $c'$ of the input, and $\mathbf{SPD}^c(\omega)$ is the spectral power distribution of the light that composes channel $c$ of the input.

As noted in the main text, Eq. (S7) can be simplified into the form:

$$I_{out}(x, y) = \sum_c^{C_{in}} \left[ \sum_{c'}^{C_{in}} A_{c'c} \mathbf{PSF}^{c'}(\eta, \xi) \right] \star I^c_{in}(\eta, \xi), \quad (S8)$$

where $A_{c'c} = \int \mathbf{SSF}(\omega)\mathbf{PCE}^{c'}(\omega)\mathbf{SPD}^c(\omega)d\omega$, and we can define a *cross-talk matrix* whose elements are $A_{c'c}$ for the different values of $c'$ and $c$.

In the ideal case, input channels are generated by optical sources whose spectral power distributions $\mathbf{SPD}^c(\omega)$ are delta functions centered at a frequency associated with each RGB channel. Additionally, in this ideal case, meta-elements associated with a given channel have a polarization conversion efficiency spectra $\mathbf{PCE}^{c'}(\omega)$ that is equal to 0 almost everywhere and equal to 1 only at the peak frequency of the spectral power distribution associated with the matching channel of the input image. Finally, in the ideal case, the detector has a sensitivity spectrum $\mathbf{SSF}(\omega)$ that is constant in the whole frequency domain of interest. Under these conditions, the intensity profile of the convolutional layers' output is proportional to the intensity profile of the captured image, which is given by:

$$I_{out}(x, y) = \sum_c \mathbf{PSF}^c(\eta, \xi) \star I_{in}^c(\eta, \xi). \tag{S9}$$

We note that Eq. (S8) has the same functional form as Eq. (S9), as we can write it as

$$I_{out}(x, y) = \sum_c B^c(\eta, \xi) \star I_{in}^c(\eta, \xi), \tag{S10}$$

where $B$ is a tensor with the same dimensions as the multi-channel PSF. This tensor $B$ is defined such that channel $c$ of $B$, denoted as $B^c$, is given by:

$$B^c = \sum_{c'} A_{c'c} \mathbf{PSF}^{c'}. \tag{S11}$$

Finally, we should note that the cross-talk matrix $A$ is reduced to an identity matrix in the ideal case, where the conditions described above hold true. When that happens, Eq. (S8) and Eq. (S10) are reduced to Eq. (S9).

The equations above are necessary for co-training the phase profile along with the trainable parameters of the suffix layers of the neural network using the backpropagation process. In order to perform the backpropagation process, we need to calculate the gradient of the network's loss function $L$ (not to be confused with $L_\Phi$ from the previous section) with respect to the phase profile, i.e. $\nabla_\Phi L$, and we need to express it in terms of the gradients of the loss function with respect to the parameters of the network's layers. In the following sections, we'll derive the expressions for $\nabla_\Phi L$ in both the ideal case (with no cross-talk) and the non-ideal case (with cross-talk). We need these to be expressed in terms of operations that can be efficiently performed between numerical arrays, using the built-in optimized algorithms of libraries such as Numpy and Tensorflow.

**Ideal case**

The math for the ideal case will serve as a baseline for that of the non-ideal case, and because of that, we'll start with the former one. In our pipeline, the phase profile array $\Phi$ (three-dimensional, with the third dimension corresponding to channels) is used to yield a $\mathbf{PSF}$ (also a three-dimensional array). Then, from the yielded $\mathbf{PSF}$ we extract the convolutional kernels $W$ of the first convolutional layer. These convolutional kernels will be the ones that are used to process the input of the network as it goes through the first convolutional layer. As such, when the network is trained using a library such as Tensorflow via the backpropagation process, the program will have to calculate the gradient of the loss function with respect to $W$, i.e. $\nabla_W L$.

Under this scheme, $L$ depends on $W$, which in turn depends on $\mathbf{PSF}$, which itself depends on $\Phi$. Because of this, due to the chain rule of derivatives, the gradient of $L$ with respect to $\Phi$ can be expressed as:

$$\nabla_\Phi L = \sum_{\mathbf{PSF}} \frac{\partial L}{\partial \mathbf{PSF}} \nabla_\Phi \mathbf{PSF}, \tag{S12}$$

where $\frac{\partial L}{\partial \mathbf{PSF}}$ are the elements of $\nabla_{\mathbf{PSF}} L$, which can be expressed as:

$$\nabla_{\mathbf{PSF}} L = \sum_W \frac{\partial L}{\partial W} \nabla_{\mathbf{PSF}} W, \tag{S13}$$

where $\frac{\partial L}{\partial W}$ are the elements of $\nabla_W L$, which is calculated by the program's backpropagation algorithm. Because of the above, in order to calculate $\nabla_\Phi L$, we first need to find a way to express and efficiently compute $\nabla_{\mathbf{PSF}} L$ and the $\nabla_\Phi \mathbf{PSF}$.

Let's begin by finding a $\nabla_{\mathbf{PSF}}L$. Since each element of the convolutional kernel tensor $W$ is obtained by adding and subtracting specific elements of the **PSF** tensor, each of the $\nabla_{\mathbf{PSF}}W$ tensors in the sum in Eq. (S13) will be sparse. For a given $W$ (here, $W$ denotes an element of the convolutional kernel tensor), the corresponding $\nabla_{\mathbf{PSF}}W$ will be a tensor with the same dimensions as **PSF** that will be non-zero only in the index positions of the elements of the **PSF** tensor that are involved in the computation of that element $W$ of the conv-weight tensor. The non-zero values will be either 1 or $-1$. If during the phase optimization process $\Phi$ was upscaled by a scale factor $n$ (with respect to the target PSF array that is used to encode the weights obtained from the training pipeline's Step 1), then two tiles of $n \times n \times 1$ pixels of **PSF** are used to calculate one element of tensor $W$. One tile is subtracted from the other, and then the result is averaged to yield the numerical value of element $W$. Because of this, for a given element $W$, the corresponding $\nabla_{\mathbf{PSF}}W$ will be a sparse array where there will be a tile of $n \times n \times 1$ pixels with values $+\frac{1}{n^2}$ and another such tile with values $-\frac{1}{n^2}$. As such, each term $\frac{\partial L}{\partial W}\nabla_{\mathbf{PSF}}W$ in the sum of Eq. (S13) will simply be an sparse array with the same dimensions as **PSF** that contains a tile of $n \times n \times 1$ pixels whose values are all $+\frac{1}{n^2}\frac{\partial L}{\partial W}$ and another such tile whose values are all $-\frac{1}{n^2}\frac{\partial L}{\partial W}$. This way, when performing a sum over elements $W$ of all this terms, the result $\nabla_{\mathbf{PSF}}L$ will simply be a tiled version of $\nabla_W L$. As such, we have now found a way to efficiently compute $\nabla_{\mathbf{PSF}}L$ in terms of $\nabla_W L$: We just need to take $\nabla_W L$ and perform a tiling process similar to the one that we use during step 2 of the training pipeline, where we obtain a target PSF by tiling the convolutional kernel tensor $W$ we obtained from step 1.

Now, let's look at how to calculate the $\nabla_\Phi \mathbf{PSF}$ that appear in Eq. (S12), i.e. the gradients of each pixel in the **PSF** array with respect to the phase profile. First, let's look at the expression for element $(k, l, c)$ of the **PSF** array, taking into account how it depends on $\Phi$ as $\mathbf{PSF}(\Phi) = \left|\mathcal{F}^{-1}\left\{e^{i\Phi}\right\}\right|^2$:

$$\mathbf{PSF}_{k,l,c} = \left|\sum_{m=0}^{M-1}\sum_{n=0}^{N-1} \exp\left(i\Phi_{mn}^c\right) \exp\left(-2\pi i\left(\frac{mk}{M} + \frac{nl}{N}\right)\right)\right|^2, \quad (S14)$$

where $M$ and $N$ are the number of pixels in the vertical (rows) and horizontal (columns) spatial dimensions in the $\Phi$ array, respectively. Given this, we can now calculate $\frac{\partial \mathbf{PSF}_{klc}}{\partial \Phi_{m'n'}^{c'}}$:

$$\frac{\partial \mathbf{PSF}_{klc}}{\partial \Phi_{m'n'}^{c'}} = \left(\sum_{m=0}^{M-1}\sum_{n=0}^{N-1} \exp\left(i\Phi_{mn}^c\right) \exp\left(-2\pi i\left(\frac{mk}{M} + \frac{nl}{N}\right)\right)\right) \frac{\partial}{\partial \Phi_{m'n'}^{c'}} \left(\sum_{m=0}^{M-1}\sum_{n=0}^{N-1} \exp\left(-i\Phi_{mn}^c\right) \exp\left(+2\pi i\left(\frac{mk}{M} + \frac{nl}{N}\right)\right)\right)$$

$$+ \left(\sum_{m=0}^{M-1}\sum_{n=0}^{N-1} \exp\left(-i\Phi_{mn}^c\right) \exp\left(+2\pi i\left(\frac{mk}{M} + \frac{nl}{N}\right)\right)\right) \frac{\partial}{\partial \Phi_{m'n'}^{c'}} \left(\sum_{m=0}^{M-1}\sum_{n=0}^{N-1} \exp\left(i\Phi_{mn}^c\right) \exp\left(-2\pi i\left(\frac{mk}{M} + \frac{nl}{N}\right)\right)\right)$$

$$\frac{\partial \mathbf{PSF}_{klc}}{\partial \Phi_{m'n'}^{c'}} = -i\delta_{cc'} \left(\mathcal{F}^{-1}\left\{\exp\left(i\Phi^c\right)\right\}_{kl}\right) \left(\exp\left(i\Phi_{m'n'}^{c'} - 2\pi i\left(\frac{m'k}{M} + \frac{n'l}{N}\right)\right)\right)^*$$

$$+ i\delta_{cc'} \left(\mathcal{F}^{-1}\left\{\exp\left(i\Phi^c\right)\right\}_{kl}\right)^* \left(\exp\left(i\Phi_{m'n'}^{c'} - 2\pi i\left(\frac{m'k}{M} + \frac{n'l}{N}\right)\right)\right),$$

(S15)

where $\delta_{cc'}$ is a Kronecker delta between channel index values $c$ and $c'$ and the $*$ asterisk denotes complex conjugation. The expression in Eq. (S15) can be simplified if we take into account that (it can be shown that) for any two complex numbers $x$ and $y$, we have that $-xy^* + x^*y = -2i\text{Im}\left[xy^*\right]$, where Im denotes imaginary part of a complex number. This way, we finally have:

$$\frac{\partial \mathbf{PSF}_{klc}}{\partial \Phi_{mn}^{c'}} = 2\text{Im}\left[\exp\left(-i\Phi_{mn}^{c'}\right) \mathcal{F}^{-1}\left\{\exp\left(i\Phi^c\right)\right\}_{kl} \exp\left(2\pi i\left(\frac{mk}{M} + \frac{nl}{N}\right)\right)\right] \delta_{cc'}. \quad (S16)$$

4

With this, we can calculate the $(m, n, c')$ element of $\nabla_\Phi L$ using Eq. (S12) and Eq. (S16) as:

$$
\begin{aligned}
(\nabla_\Phi L)_{mnc'} &= \sum_{k,l,c} \frac{\partial L}{\partial \mathbf{PSF}_{klc}} \frac{\partial \mathbf{PSF}_{klc}}{\partial \Phi^{c'}_{mn}} \\
(\nabla_\Phi L)_{mnc'} &= \sum_{k,l,c} \frac{\partial L}{\partial \mathbf{PSF}_{klc}} 2\mathrm{Im}\left[\exp\left(-i\Phi^{c'}_{mn}\right) \mathcal{F}^{-1}\left\{\exp\left(i\Phi^c\right)\right\}_{kl} \exp\left(2\pi i\left(\frac{mk}{M} + \frac{nl}{N}\right)\right)\right] \delta_{cc'} \\
(\nabla_\Phi L)_{mnc'} &= 2\mathrm{Im}\left[\sum_{k,l} \frac{\partial L}{\partial \mathbf{PSF}_{klc'}} \exp\left(-i\Phi^{c'}_{mn}\right) \mathcal{F}^{-1}\left\{\exp\left(i\Phi^{c'}\right)\right\}_{kl} \exp\left(2\pi i\left(\frac{mk}{M} + \frac{nl}{N}\right)\right)\right] \\
(\nabla_\Phi L)_{mnc'} &= 2\mathrm{Im}\left[(\exp(-i\Phi))_{mnc'} \sum_{k,l} \frac{\partial L}{\partial \mathbf{PSF}_{klc'}} \mathcal{F}^{-1}_{2\mathrm{D}}\left\{\exp(i\Phi)\right\}_{klc'} \exp\left(2\pi i\left(\frac{mk}{M} + \frac{nl}{N}\right)\right)\right] \\
(\nabla_\Phi L)_{mnc'} &= 2\mathrm{Im}\left[(\exp(-i\Phi))_{mnc'} \sum_{k,l} \left((\nabla_{\mathbf{PSF}} L)\left(\mathcal{F}^{-1}_{2\mathrm{D}}\left\{\exp(i\Phi)\right\}\right)\right)_{klc'} \exp\left(2\pi i\left(\frac{mk}{M} + \frac{nl}{N}\right)\right)\right] \\
(\nabla_\Phi L)_{mnc'} &= 2\mathrm{Im}\left[(\exp(-i\Phi))_{mnc'} \mathcal{F}_{2\mathrm{D}}\left\{(\nabla_{\mathbf{PSF}} L)\left(\mathcal{F}^{-1}_{2\mathrm{D}}\left\{\exp(i\Phi)\right\}\right)\right\}_{mnc'}\right],
\end{aligned}
\tag{S17}
$$

thus, we can finally calculate $\nabla_\Phi L$ in terms of operations between known numerical arrays as:

$$
\nabla_\Phi L = 2\mathrm{Im}\left[e^{-i\Phi} \mathcal{F}_{2\mathrm{D}}\left\{(\nabla_{\mathbf{PSF}} L)\left(\mathcal{F}^{-1}_{2\mathrm{D}}\left\{e^{i\Phi}\right\}\right)\right\}\right] \tag{S18}
$$

**Non-ideal case**

The non-ideal case is very similar to the ideal one. However, the main difference mathematically is that we don't extract the elements of $W$ from the $\mathbf{PSF}$ tensor, but from the $B$ tensor defined in Eq. (S11) instead. As such, we can write:

$$
\nabla_\Phi L = \sum_B \frac{\partial L}{\partial B} \nabla_\Phi B, \tag{S19}
$$

where $\frac{\partial L}{\partial B}$ are the elements of $\nabla_B L$, which can be expressed as:

$$
\nabla_B L = \sum_W \frac{\partial L}{\partial W} \nabla_B W, \tag{S20}
$$

where $\frac{\partial L}{\partial W}$ are the elements of $\nabla_W L$, which is calculated by the program's backpropagation algorithm.

In the non-ideal case, $\nabla_B L$ can be calculated in the same way as $\nabla_{\mathbf{PSF}} L$ was calculated in the ideal case. This is because in the non-ideal case, the convolutional kernel tensor $W$ is built from the $B$ tensor in the same way as $W$ was built from the $\mathbf{PSF}$ tensor in the ideal case. Because of this, we already know what the $\frac{\partial L}{\partial B}$ that appear in the sum of Eq. (S19) look like, and we know how to efficiently compute $\nabla_B L$. As such, we can now proceed to find an expression for the $(m, n, c')$

element of $\nabla_\Phi L$. We have:

$$(\nabla_\Phi L)_{mnc'} = \sum_{k,l,c} \frac{\partial L}{\partial B_{klc}} \frac{\partial B_{klc}}{\partial \Phi^{c'}_{mn}}$$

$$(\nabla_\Phi L)_{mnc'} = \sum_{k,l,c} \frac{\partial L}{\partial B_{klc}} \frac{\partial}{\partial \Phi^{c'}_{mn}} \left( \sum_{c''} A_{c''c} \mathbf{PSF}_{klc''} \right)$$

$$(\nabla_\Phi L)_{mnc'} = \sum_{k,l,c} \frac{\partial L}{\partial B_{klc}} \sum_{c''} A_{c''c} \frac{\partial \mathbf{PSF}_{klc''}}{\partial \Phi^{c'}_{mn}}$$

$$(\nabla_\Phi L)_{mnc'} = \sum_{k,l,c} \frac{\partial L}{\partial B_{klc}} A_{c'c} \frac{\partial \mathbf{PSF}_{klc'}}{\partial \Phi^{c'}_{mn}}$$

$$(\nabla_\Phi L)_{mnc'} = \sum_{k,l,c} \frac{\partial L}{\partial B_{klc}} A_{c'c} 2\mathrm{Im}\left[ \left(e^{-i\Phi}\right)_{mnc'} \mathcal{F}^{-1}_{2D}\left\{e^{i\Phi}\right\}_{klc'} \exp\left(2\pi i \left(\frac{mk}{M} + \frac{nl}{N}\right)\right) \right]$$

$$(\nabla_\Phi L)_{mnc'} = 2\mathrm{Im}\left[ \sum_{k,l,c} A_{c'c} (\nabla_B L)_{klc} \left(e^{-i\Phi}\right)_{mnc'} \mathcal{F}^{-1}_{2D}\left\{e^{i\Phi}\right\}_{klc'} \exp\left(2\pi i \left(\frac{mk}{M} + \frac{nl}{N}\right)\right) \right]$$

$$(\nabla_\Phi L)_{mnc'} = 2\mathrm{Im}\left[ \left(e^{-i\Phi}\right)_{mnc'} \sum_{k,l} \left(\mathcal{F}^{-1}_{2D}\left\{e^{i\Phi}\right\}_{klc'} \exp\left(2\pi i \left(\frac{mk}{M} + \frac{nl}{N}\right)\right)\right) \left(\sum_c A_{c'c}(\nabla_B L)_{klc}\right) \right]. \tag{S21}$$

Let's define a tensor $H$ such that $H^{c'}_{kl} = \sum_c A_{c'c}(\nabla_B L)_{klc}$. With that, we can further simplify Eq. (S21) as:

$$(\nabla_\Phi L)_{mnc'} = 2\mathrm{Im}\left[ \left(e^{-i\Phi}\right)_{mnc'} \sum_{k,l} \left(\mathcal{F}^{-1}_{2D}\left\{e^{i\Phi}\right\} H\right)_{klc'} \exp\left(2\pi i \left(\frac{mk}{M} + \frac{nl}{N}\right)\right) \right],$$

$$(\nabla_\Phi L)_{mnc'} = 2\mathrm{Im}\left[ \left(e^{-i\Phi}\right)_{mnc'} \mathcal{F}_{2D}\left\{H \mathcal{F}^{-1}_{2D}\left\{e^{i\Phi}\right\}\right\}_{mnc'} \right] \tag{S22}$$

thus, we can finally calculate $\nabla_\Phi L$ in terms of operations between numerical arrays as:

$$\nabla_\Phi L = 2\mathrm{Im}\left[ e^{-i\Phi} \mathcal{F}_{2D}\left\{H \mathcal{F}^{-1}_{2D}\left\{e^{i\Phi}\right\}\right\} \right], \tag{S23}$$

where $H$ is a tensor defined such that its channel $c'$, denoted $H^{c'}$, is given by $H^{c'} = \sum_c A_{c'c}(\nabla_B L)^c$, with $(\nabla_B L)^c$ being channel $c$ of $\nabla_B L$.



\end{document}